\pdfoutput=1
\documentclass[11pt]{article}

\usepackage[final]{acl}

\usepackage{times}
\usepackage{latexsym}

\usepackage[T1]{fontenc}
\usepackage[utf8]{inputenc}

\usepackage{microtype}

\usepackage{inconsolata}

\usepackage{graphicx}

\usepackage{graphicx}
\usepackage{tabularx}
\usepackage{amsmath}
\usepackage{amssymb}
\usepackage{multirow}
\usepackage{makecell}
\usepackage{spverbatim}
\usepackage{fancyvrb,xcolor}
\usepackage{csquotes}
\usepackage{tcolorbox}
\tcbuselibrary{listingsutf8}
\usepackage{booktabs}
\newcolumntype{Y}{>{\centering\arraybackslash}X}
\usepackage{subcaption}
\usepackage{enumitem}
\title{\textsc{SolAr}: Towards Characterizing Subjectivity of Individuals through \\Modeling Value Conflicts and Trade-offs}

\author{Younghun Lee \and Dan Goldwasser \\
       Department of Computer Science\\
        Purdue University\\
        West Lafayette, IN, USA\\
        \texttt{\{younghun, dgoldwas\}@purdue.edu} \\}

\begin{document}
\maketitle
\begin{abstract}
Large Language Models (LLMs) not only have solved complex reasoning problems but also exhibit remarkable performance in tasks that require subjective decision-making. Existing studies suggest that LLM generations can convey subjectivity to some extent, yet exploring whether LLMs can account for individual-level subjectivity has not been sufficiently studied. In this paper, we characterize the subjectivity of individuals on social media and infer their moral judgments using LLMs. We propose a framework, \textsc{SolAr} (\textsc{\textbf{S}}ubjective Gr\textsc{\textbf{o}}und with Va\textsc{\textbf{l}}ue \textsc{\textbf{A}}bst\textsc{\textbf{r}}action), that observes value conflicts and trade-offs in the user-generated texts to better represent subjective ground of individuals. Empirical results demonstrate that our framework enhances overall inference performance, with notable improvements for users with limited data and in controversial situations. Additionally, we qualitatively show that \textsc{SolAr} provides explanations about individuals' value preferences, which can further account for their judgments.
\end{abstract}

\section{Introduction}
For the last few years, Large Language Models (LLMs) have shifted the paradigm of solving NLP problems to autoregressive language generation and achieved human-like performance in many downstream tasks \citep{2020t5, wei2022chain, kojima2022large, wang2022self, yu2022generate, he2024can}. Not only have LLMs solved objective problems that require complex reasoning skills, such as STEM-related questions \citep{imani2023mathprompter, wang2024mmlu, abbasiantaeb2024let}, they also exhibit remarkable performance in subjective decision-making processes, such as detecting toxicity \citep{hartvigsen2022toxigen}, generating model evaluation \citep{perez2022discovering}, following ethical principles \citep{bai2022constitutional}, etc.

Recent studies explore whether LLMs can generate perspectives and reasoning that align well with a specific persona or demographic information \citep{durmus2023towards, nie2024moca, zheng2024helpful}. 
The results of these studies show that it is possible, to an extent, to ground LLM generations with these traits, however, LLMs tend to rely on superficial facts and assumptions about the roles and demographic traits rather than apply a deeper understanding.

Our goal in this paper is to study a different aspect of subjectivity, focusing on the \textit{individual-level} (rather than generalizing over demographic traits), which has not been sufficiently studied yet. As personalized AI becomes more widely used \citep{mcclain2024chatgpt}, understanding whether LLMs can be utilized to characterize individual subjectivity becomes more important.
Analyzing individual-level subjectivity using LLMs faces two main challenges. The first challenge is guiding LLM generation to be consistent with a specific subjective view. Existing methods for capturing subjectivity at the level of a generalized persona, role, or demographic information show that it is not trivial to steer LLMs to follow certain aspects \citep{durmus2023towards, zheng2024helpful}. Second, even if an optimal approach for grounding subjective aspects in LLMs' generations existed, expressing these aspects at the level of an individual is not straightforward (i.e., two instances of the same persona could still have different subjective preferences). Conceptualizing and operationalizing subjectivity at this level poses a second challenge.

\begin{figure*}[t]
    \centering
    \includegraphics[width=0.99\textwidth]{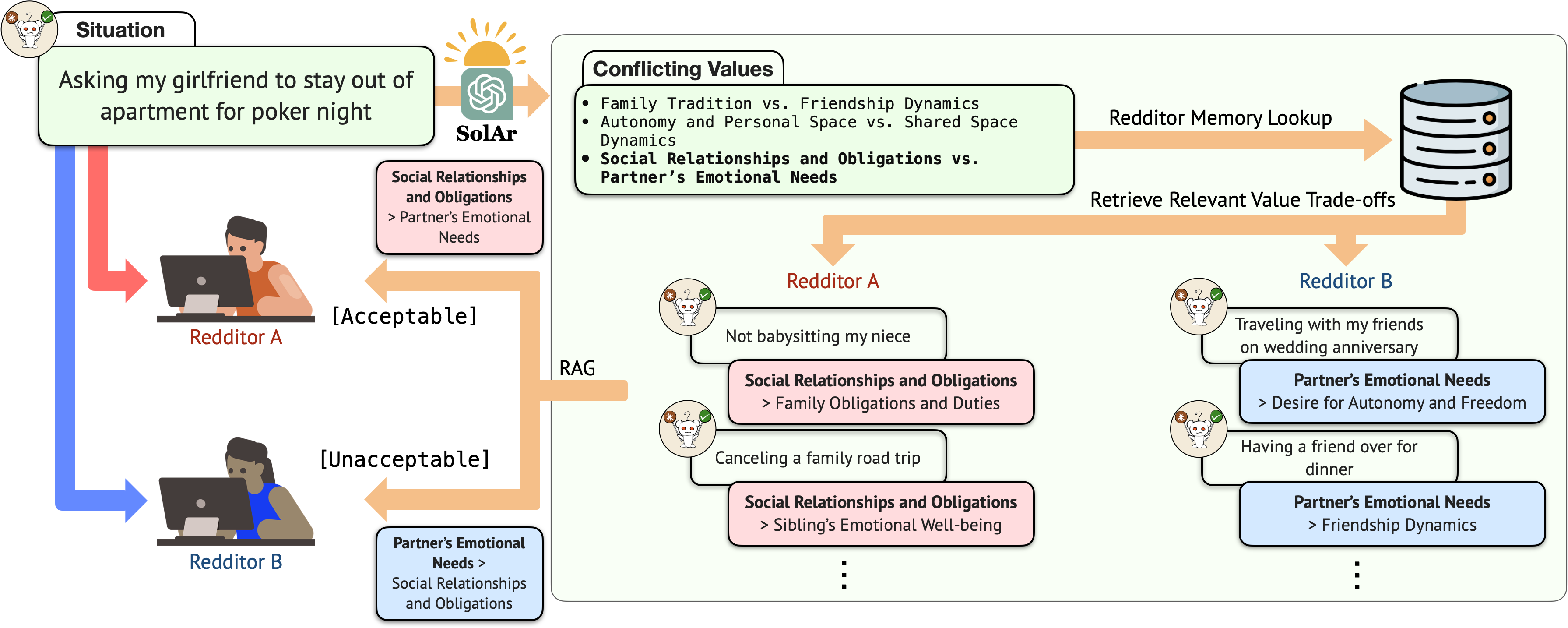}
    \caption{Inference process of our framework, \textsc{SolAr}. When an input situation is given, \textsc{SolAr} identifies conflicting values in the story, and retrieves the most relevant value trade-off history from the redditor's past comments. The retrieved subjective ground is then added to the prompt to infer each redditor's likely judgments to the situation.}
    \label{fig:model}
\end{figure*}

In this paper, our objective is to characterize the subjectivity of individuals with LLMs by analyzing users' behaviors in a Reddit community, \texttt{r/AmITheAsshole}. In this community, original posters write about situations where they have conflicts with others and ask whether their behaviors are acceptable, and other redditors leave comments with their judgments. We formulate a classification task, predicting whether a specific redditor would judge a test situation acceptable or unacceptable, for evaluating different models' ability to capture subjectivity. In order to perform the classification task more efficiently, we hypothesize that redditors' \textit{subjective ground}, principles that play a fundamental role in making moral judgments on others' situations \citep{neuhouser1990fichte}, can be represented by decomposing their past behaviors; we use redditors' past comments in the community to determine their likely judgments on unseen situations.

More specifically, we adopt value pluralism when considering redditors' past comments. Value pluralism suggests that there are multiple values that may be equally correct, yet in conflict with one another \citep{crowder1998value, galston2002liberal}. Psychological studies adopt this idea and argue that the human cognitive system makes novel judgments by making trade-offs between conflicting values when it encounters situations with colliding moral intuitions \citep{fiske1997taboo, guzman2022moral}. In this paper, we propose a framework, \textsc{SolAr} (\textsc{\textbf{S}}ubjective Gr\textsc{\textbf{o}}und with Va\textsc{\textbf{l}}ue \textsc{\textbf{A}}bst\textsc{\textbf{r}}action), that observes trade-offs between conflicting values in the user-generated texts (i.e. comments), identifies the most relevant value conflicts with respect to the input situation, and infer the redditors' likely judgments using off-the-shelf LLMs. Our framework aims to tackle the two challenges of characterizing individual-level subjectivity described above; it conceptualizes individual-level subjectivity with value trade-offs, and grounds LLM generations with Retrieval-Augmented Generation (RAG). Empirical results show that \textsc{SolAr} distinguishes redditors' subjective preferences and further provides explanations of their value trade-offs. Our framework improves the overall performance of downstream tasks and better guides LLMs, especially in low-resource user settings (i.e. users with a limited amount of data) and morally controversial scenarios. Figure \ref{fig:model} shows an example of how \textsc{SolAr} determines the judgment of different redditors given a situation.

\noindent\textbf{Key Contributions}: To the best of our knowledge, this is the first attempt to explore whether off-the-shelf LLMs can be effectively used to account for individual-level subjectivity in the real-world online community. With value abstraction, we highlight that redditors show distinct subjectivity patterns. Additionally, we propose a novel framework that encompasses a value trade-off system and performs better in downstream tasks as well as grounding LLM generations.

\section{Problem Formulation}
\label{sec:prob-formulation}
In this section, we describe four major elements that formulate modules, learning processes, and inference tasks.

\noindent\textbf{Situation}\quad Situation refers to the text description of what has happened in the real world. We focus on situations that portray conflicts so that one's point of view can be projected in diverse ways. \textit{``Asking my girlfriend to stay out of the apartment for poker night''} is an example of a situation.

\noindent\textbf{Individual}\quad This research aims to analyze different individuals' reactions and responses to various social situations. Unlike opinion mining or social commonsense reasoning studies, where perspectives and rationales are represented as an aggregate of a large number of humans, our main focus is to observe distinct patterns that characterize each individual and understand the rationales behind their decision-making processes. 
% We hypothesize that these patterns can be captured by observing how individuals react to the situations that are controversial. In this research, we consider moral judgments on the situations (i.e. whether the situation is morally acceptable or not) to account for individuals.

\noindent\textbf{Subjective Ground}\quad Subjective Ground refers to principles or maxims that steer individuals' moral judgments or perspectives. \textit{``One should put their significant other's needs as a top priority.''} is a subjective ground item that is relevant to the example situation above. Every individual has a distinct subjective ground and is built from various aspects such as their past experience, demographics, personality, etc. \citep{eysenck1975manual,schwaba2023subjective, schoeller2024predicting}. One of our research hypotheses is that the subjective ground of individuals can be inferred by observing their past behaviors. Throughout this research, we aim to validate this hypothesis by modeling the subjective ground with individuals' comment history on social media.

\noindent\textbf{Value Abstraction}\quad One of the limitations of the aforementioned assumption is that it works in a setting in which individuals' past behaviors can be observed across an extremely large range of social situations---an impractical scenario in real-world contexts. Thus in reality, when inferring individuals' likely judgments on unseen situations, it is necessary to formulate some hypotheses based on observable past history. Values, which ``guide the selection or evaluation of actions, policies, people, and events'', can serve to induce hypotheses about individuals' moral judgments \citep{schwartz2012overview}. Value abstraction refers to a process that maps past history (i.e. comments) to a high-level representation of subjectivity that can be generalized to a broader spectrum of situations (i.e. values). 
% In this paper, we explore several approaches for abstracting observed.

\section{Task Description}
We analyze subjective perspectives and judgments of individuals posted on a Reddit community, \texttt{r/AmITheAsshole}. In this community, original posters (OP) describe situations in which they have conflicts with others and ask if they are at fault. Other individuals (i.e. redditors) then leave comments and judge the acceptability of the OP's behaviors in the situation. Redditors' judgments are pre-coded words in the comment; \verb|YTA| (You’re The Asshole), \verb|YWBTA| (You Would Be The Asshole), \verb|NTA| (Not The Asshole), \verb|YWNBTA| (You Would Not Be The Asshole), \verb|ESH| (Everyone Sucks Here), \verb|NAH| (No Assholes Here), and \verb|INFO| (Not Enough Info). For simplicity, we group \verb|NTA|, \verb|NAH|, and \verb|YWNBTA| as `acceptable', and
\verb|YTA|, \verb|ESH|, and \verb|YWBTA| as `unacceptable'. \verb|INFO| is discarded as it does not convey subjectivity. We aim to learn the subjectivity of individuals by analyzing redditors' comments and judgment patterns on various situations.

There are a couple of benefits to using this community in analyzing individual-level subjectivity with language models. First, the situations described in this community are mostly about everyday events that are generic (e.g. \textit{``not attending a friend's wedding''}) rather than related to specific world events (e.g. \textit{``commenting on the new executive orders from the president''}); thus, the language models can have a better understanding of the situations without having a knowledge gap. Another benefit is that the redditors' subjective judgments are coded in a discrete fashion, which makes language model predictions more objective compared to open-ended analysis of subjectivity.

\begin{table}\footnotesize
\centering
\begin{tabularx}{0.98\columnwidth}{l|r}
\hline
\multicolumn{2}{c}{Truncated \texttt{r/AmITheAsshole}}\\
\hline
\# of total instances &  53,280\\
\# of unique situations &  17,432\\
\# of unique redditors &  100\\
Max / Min \# of instances per redditor &  2,870 / 148 \\
\# Accept. / Unaccept. labels (overall) &  38,365 / 14,915\\
\# Accept. / Unaccept. labels (skewed) &  2,615 / 127\\
\hline
\end{tabularx}
\caption{\label{tab:dataset}
Statistics of the truncated dataset we use for training and inference. Label distributions are described in two ways; by combining all redditors (overall), and by combining three redditors who showed the most skewed judgment patterns (skewed).
}
\end{table}

\subsection{Crawling from \texttt{r/AmITheAsshole}}
\label{sec:data-crawl}
We crawl all posts in the \texttt{r/AmITheAsshole} community from November 2014 to June 2023 and filter out threaded comments to ensure all comments solely argue the acceptability of the situations. In order to perform training and inference more efficiently, we narrow it down to 1.7K unique reddit posts and 100 redditors who commented on these posts. We discuss how we truncate the data and whether the truncated situations and redditors appropriately represent the population distribution in the Appendix \ref{sec:appendix-data-dist}. Detailed statistics of the truncated dataset are described in Table \ref{tab:dataset}, and the data is released for future reproduction \footnote{https://github.com/younggns/solar}.
%We crawl all posts in the \texttt{r/AmITheAsshole} community from November 2014 to June 2023 and identify the top 8 redditors who commented the most to ensure a sufficient amount of data for each individual. To ensure that the comments solely mention the situations, we consider top-level comments without including any threaded comments. We then filter out the posts for which the top 8 users did not leave their judgments. Detailed statistics of the crawled dataset are described in Table \ref{tab:dataset}.

\subsection{Abstract Value Annotation}
As discussed in \ref{sec:prob-formulation}, we produce a high-level abstraction of redditors' comments and the situations. For each pair of situations and an individual's comments, we prompt an off-the-shelf LLM\footnote{We use OpenAI's \textit{gpt-4o-0806} model} and generate values that are observed from the situation and redditors' comments. As human values can be defined and captured differently in the same text, we apply several different approaches. 

We first analyze and annotate the texts using a top-down approach based on the theory of basic human values proposed by \citet{schwartz1992universals}. The theory suggests ten basic values that could explain how people in different cultures recognize the underlying motivation and goals. One of the advantages of using this framework is that it explains values that align or conflict with one another, which naturally expands hypotheses of individual subjectivity. %For instance, when we observe an individual's preference of \textit{``Openness to Change''}, it is implied that the individual is likely to have less preference of \textit{``Conservation''} or \textit{``Tradition''}. 
Detailed explanations of the ten basic human values and how they are annotated are described in Appendix \ref{sec:schwartz-theory}.

As opposed to using a fixed set of values for characterizing subjectivity, we use a bottom-up approach to discover more open-ended values observed in the texts. In this setup, we apply value trade-off theories. Humans make judgments based on value trade-offs when different values conflict to each other \citep{fiske1997taboo, leyva2019towards, guzman2022moral}. As situations in the dataset mostly describe conflicts OPs are having, we prompt LLMs to identify all conflicting value pairs in the situation and discover value trade-offs made by each redditor in the comments.

When conflicting values are generated by LLMs, the level of abstraction is insufficient; values describe situation-specific details (e.g. \textit{``prioritizing girlfriend's plan to cook together''}), rather than general concepts (e.g. \textit{``partner's emotional needs''}). To address this, we iteratively cluster similar value representations and discover high-level definitions for these clustered values. We follow the approach used in \citet{lam2024concept} to derive abstract value representations. Detailed processes for prompting LLMs to generate value trade-offs and clustering values are described in Appendix \ref{sec:cluster-value-conflicts}. With more nuanced representations, we argue that the generated values provide richer contextual information than Schwartz's values and ultimately help better characterize individual subjectivity. As demonstrated in Appendix \ref{sec:appendix-value-comparison}, the values produced by the LLM integrate multiple Schwartz values, reflecting the multi-dimensional nature of subjective preferences.

\subsection{Learning Problem}
Our main focus is to use language models as a reasoning agent to understand individual-level subjectivity and predict their likely behaviors in unseen instances. More specifically, we formulate a binary classification task where the language models are given situations and redditors' most relevant subjective ground, either their past comments or value preference, and predict whether the redditors would judge the OP's behaviors acceptable or not. 

\section{Model}
In this research, we propose a framework, \textsc{SolAr}, that accomplishes the task with Retrieval-Augmented Generations (RAG). 

Let $\mathcal{U} = \{u_1, u_2, \dots, u_n\}$ denote the set of all distinct redditors, and $\mathcal{S} = \{s_1, s_2, \dots, s_N\}$ indicate the set of situations (i.e. reddit posts). When an $i$-th redditor commented on a $j$-th situation, we define their comment as $c_{ij}$ and the acceptability judgment as $y_{ij}$ where $y_{ij} \in \{0, 1\}$. More specifically, $\mathcal{C}=\{c_{ij} \ |\ u_i \ \text{commented on} \ s_j\}$ and $\mathcal{Y}=\{ y_{ij} \ | \ u_i \ \text{commented on} \ s_j\}$. After we annotate the moral values of the situations and comments, we get the values of each situation as $s_{j}^{\mathcal{V}}$ and the values of each comment as $c_{ij}^\mathcal{V}$. Using a dynamic embedding model $f_{embed}$, all situations, comments, and values are transformed into vectors\footnote{We use OpenAI's \textit{text-embedding-3-large} model.}. We denote these vectors by bold letters such as $\mathbf{s}_j = f_{embed}(s_j)$.

\subsection{Subjective Ground Retrieval}
\label{sec:subjective-ground-retrieval}
We first define the redditor history data where the retrieval function searches for the most relevant instances to the input. We keep data separate for each redditor and let $\mathcal{D}_{i}$ denote the history data of an $i$-th redditor. $\mathcal{D}_{i}$ contains representations of the situations and comments, namely $\mathcal{D}_{i}=\{ (s_j, s_j^\mathcal{V}, c_{ij}, c_{ij}^\mathcal{V}, y_{ij}) \ |\ u_i \ \text{commented on} \ s_j\}$.

In order to infer how the target individual would react to a test situation, $x$, we design several heuristics to retrieve the most relevant instances. First, we query the retrieval function with the raw situation representations. Each query consists of a user indicator and a vector representation of a test situation, $\mathbf{x}$. Then the retrieval function outputs $k$ nearest instances with respect to the Euclidean distance between the test situation and situations in the history data $\mathcal{D}$. Let $R( \ )$ be the retrieval function, we define subjective ground retrieval with situations by:

\begin{equation}\label{eqn1}
R(u_i, \mathbf{x})=\underset{\mathbf{s}_j \in \mathcal{D}_i}{\mathrm{top}\text{-}k} \; \text{dist}(\mathbf{x}, \mathbf{s}_j)^{-1}    
\end{equation}

where $\text{dist}(\mathbf{x}, \mathbf{s}_j)=\|\mathbf{x} - \mathbf{s}_j\|_2$

%subjective ground needs to be retrieved. 
% For example, the subjective ground example, \textit{``prioritizing girlfriend's plan to cook together''}, would be relevant to the situation, \textit{``asking my girlfriend to stay out of apartment for poker night''}, but it would be irrelevant to other situations related to wedding ceremonies.
%We design several heuristics to operationalize the retrieval, mainly by differentiating queries.

% First, we could simply select the comments that are left on situations similar to the test situations by computing the pairwise distances between the vector representations of the situations. 

% While retrieving comments from similar situations, we add another heuristics to explicitly retrieve comments that show different judgment patterns and guarantee that LLMs are prompted with different moral judgments. Suppose that LLMs are prompted to predict whether a redditor judges \textit{``not paying for my daughter's wedding''} acceptable or not. In observing the past history, it is possible that this redditor commented to the top 5 most similar situations as ``not acceptable''. We then try to find other instances where the redditor says differently. When the redditor judges as ``acceptable'' to \textit{``refusing to contribute to my daughter's wedding after she cancelled the previous one''}, LLMs not only understand the redditor's general judgment tendencies, but also observe special conditions that would change their usual judgment patterns.

As an alternative to computing the distance between situation representations, we could use abstract values. The motivation for using value representations is to retrieve comments from more diverse cases; distance among situations yields topically similar situations only, while using values can retrieve situations that are relevant in terms of high-level values. Consider, for example, an input situation about wedding ceremonies. While retrieving relevant items with situations provides wedding related instances, using abstract value representations could retrieve situations about social appearance, money, family relationship dynamics, etc. Let $R_{val}( \ )$ be the retrieval function that takes a user indicator and a value representation of a test situation, $\mathbf{x}^\mathcal{V}$, we define subjective ground retrieval with values by:

\begin{equation}\label{eqn2}
    R_{val}(u_i, \mathbf{x}^\mathcal{V})=\underset{\mathbf{s}_j^\mathcal{V} \in \mathcal{D}_i}{\mathrm{top}\text{-}k} \; \text{dist}(\mathbf{x}^\mathcal{V}, \mathbf{s}_j^\mathcal{V})^{-1}
\end{equation}

% \subsection{Adding Intermediate Reasoning}
% Reasoner module accomplishes ``filling the gap'' before performing inference task with retrieved subjective ground. Retrieved subjective ground might not provide sufficient information about the individual in order to predict their judgment on the test situation. This is because of sub-optimal performance of the retriever, or because there's no relevant evidence to the test situation in the individual's past comment history. It is also possible that the subjective ground of an individual has different preference patterns that needs to be resolved. For instance, when an individual prefers ``helping your neighbors'' over focusing on their own issues most of the time, this value selection could be entirely flipped if their issue is about their children. Reasoner module tries make hypotheses with the given subjective ground about the most likely judgments. This could be accomplished by simply prompting LLMs to fill the information gap, or fine-tune small language models to generate valid reasoning.

% \subsection{Judgment Prediction}
\subsection{Language Model Prompting}
After the retrieval functions provide the most relevant past history of a target redditor to the test situation, we prompt off-the-shelf LLMs to predict the likely judgment of the redditor. Essentially, the LLMs are considered as a general reasoner that accounts for individual subjectivity; LLMs are not trained to represent each individual's subjectivity, but they perform inference based on the given pieces of specific redditor's past subjective behaviors.

We add $k$ nearest instances as few-shot examples. In each of the few-shot examples, a short description of the retrieved situation, a redditor's comment, and their judgment are included. In order to observe the usefulness of the annotated values, we also experiment with adding values that are perceived in the comments in few-shot examples. An example prompt is described below:

\begin{tcolorbox}[title=Prompt Example with Comments Only, colback=blue!5, colframe=blue!50, fonttitle=\bfseries\footnotesize, fontupper=\footnotesize]
You will be given examples of a situation, Person X's comment on the situation, and Person X's judgment on the situation (i.e. whether it is acceptable or not).
\newline\newline
[Situation] Not babysitting my niece\newline
[Comment] When you talk to your brother ..\newline
[Judgment] Acceptable\newline
..
\newline\newline
Now you will be given a new situation. Based on your understanding of Person X's judgments on different situations, tell me how Person X would judge the new situation.\newline
[Situation] Canceling a family road trip
\end{tcolorbox}
% Given an input and a set of subjective ground provided by the retrieval process, off-the-shelf LLMs predict the likely moral judgment of a redditor. Prediction language models could be designed as a universal language model; this implies that the difference among individuals mostly comes from their subjective ground, and the output reactions derived from the retrieved subjective ground need to be reasonable in general. 
% For instance, when the predictor gets a text, \textit{``Pet owners should be responsible for the well-being of the pets''}, as a subjective ground of an individual, then it needs to predict that the individual would likely to say \textit{``Take one's cat to a poor quality dorm room''} is a situation that is ``not acceptable''. 

\section{Experiments}
In all experiments, we use the macro F1 score as the evaluation metric due to the imbalanced label distribution. For each model, we first calculate the macro F1 score individually for each redditor, then take the unweighted average of these scores across all redditors. This approach ensures that every redditor has equal influence on the final score, regardless of how many instances each contributed. Even if a model performs well on some redditors who are easily predictable, its overall score becomes lower if it fails to learn the subjectivity of other redditors.

% We also report macro F1 scores of the two specific redditors who present extremely skewed label distributions. As shown in Table \ref{tab:dataset}, these two users judge situations as ``unacceptable'' less than 8$\%$ of the time in the dataset. We consider this as another useful metric to evaluate language models' ability to have deeper understandings of subjectivity. If language models rely on superficial patterns and associations from the past history rather than reasoning about the subjective ground of the redditors, their performances are likely to be negatively affected by such skewed distributions. 

\begin{table}[t]
\centering
\small
\begin{tabularx}{\linewidth}{lYY}
\toprule
\multicolumn{3}{c}{\textbf{Moral Judgment Prediction - Baselines}} \\
\midrule
\textbf{Model} & \textbf{All Redditors} & \textbf{Top 8 Redditors} \\
\midrule
\multicolumn{3}{c}{\textit{Encoder-only Models}} \\
\midrule
DistilBERT-base-uncased     & 47.90 $^{\pm \text{0.0053}}$ & 68.71 $^{\pm \text{0.0145}}$ \\
RoBERTa-base        & 46.68 $^{\pm \text{0.0038}}$ & 70.48 $^{\pm \text{0.0088}}$ \\
DeBERTa-v3-large     & 42.45 $^{\pm \text{0.0013}}$ & 70.94 $^{\pm \text{0.0024}}$ \\
\midrule
\multicolumn{3}{c}{\textit{Encoder-Decoder Models}} \\
\midrule
BART-base     & 48.90 $^{\pm \text{0.0055}}$ & 68.50 $^{\pm \text{0.0150}}$ \\
FLAN-T5-base        & 46.21 $^{\pm \text{0.0035}}$ & 65.44 $^{\pm \text{0.0118}}$ \\
\midrule
\multicolumn{3}{c}{\textit{Encoder-Decoder Models; Seq2Seq}} \\
\midrule
BART-base     & 40.83 $^{\pm \text{0.0012}}$ & 66.67 $^{\pm \text{0.0086}}$ \\
FLAN-T5-base        & 43.01 $^{\pm \text{0.0020}}$ & 66.86 $^{\pm \text{0.0087}}$ \\
\bottomrule
\end{tabularx}
\caption{Macro F1 scores of baseline models. The average performance of fine-tuned models over all 100 redditors shows a significance drop compared to the performance of the top 8 redditors who commented over 1,000 times.}
\label{tab:experimental-results-baseline}
\end{table}

\begin{table*}[t]\footnotesize
\centering
\resizebox{\textwidth}{!}{
\begin{tabular}{c|c|cc|cc|cc}
\toprule
\multicolumn{8}{c}{\textbf{Moral Judgment Prediction - RAG-based Models}} \\
\midrule

\multirow{2}{*}{\textbf{\shortstack[l]{Retrieval \\ Strategy}}}  & \multirow{2}{*}{\textbf{\shortstack[l]{ \\ Input}}} & \multicolumn{2}{c|}{\textbf{All Redditors}} &  \multicolumn{2}{c|}{\textbf{Top 50\% Redditors}} & \multicolumn{2}{c}{\textbf{Bottom 50\% Redditors}} \\
& & All & Contro & All & Contro & All & Contro  \\
\midrule
\multirow{3}{*}{Situation} 
& Comment & 78.62$^{\pm \text{0.001}}$ & 78.61$^{\pm \text{0.002}}$ & 78.58$^{\pm \text{0.004}}$ & 80.48$^{\pm \text{0.001}}$ & 78.66$^{\pm \text{0.004}}$ & 76.53$^{\pm \text{0.000}}$ \\
& Comment + Trade-off & 79.70$^{\pm \text{0.001}}$ & 80.32$^{\pm \text{0.002}}$ & 78.62 $^{\pm \text{0.001}}$& 80.17 $^{\pm \text{0.002}}$& 80.78$^{\pm \text{0.002}}$ & 80.47$^{\pm \text{0.004}}$ \\
& Comment + Schwartz & 78.14 $^{\pm \text{0.000}}$& 79.11 $^{\pm \text{0.001}}$& 76.53 $^{\pm \text{0.000}}$& 79.45$^{\pm \text{0.000}}$ & 79.74 $^{\pm \text{0.003}}$& 78.73 $^{\pm \text{0.004}}$\\
\midrule
\multirow{2}{*}{Schwartz's Value} 
& Comment & 79.05 $^{\pm \text{0.003}}$& 81.38 $^{\pm \text{0.006}}$& 78.29 $^{\pm \text{0.005}}$& 80.97 $^{\pm \text{0.004}}$& 79.81 $^{\pm \text{0.002}}$& 81.82 $^{\pm \text{0.007}}$\\
& Comment + Schwartz & 77.35 $^{\pm \text{0.002}}$& 80.73 $^{\pm \text{0.008}}$& 76.47 $^{\pm \text{0.003}}$& 80.47$^{\pm \text{0.013}}$ & 78.24 $^{\pm \text{0.001}}$& 81.01 $^{\pm \text{0.002}}$\\
\midrule
\multirow{2}{*}{Value Trade-off} 
& Comment & 78.90 $^{\pm \text{0.000}}$& 80.76 $^{\pm \text{0.000}}$& 77.79 $^{\pm \text{0.003}}$& 81.14$^{\pm \text{0.007}}$ & 79.96 $^{\pm \text{0.001}}$& 80.78$^{\pm \text{0.009}}$ \\
& Comment + Trade-off & 78.33 $^{\pm \text{0.001}}$& 82.44 $^{\pm \text{0.005}}$& 77.26 $^{\pm \text{0.001}}$& 82.44 $^{\pm \text{0.005}}$& 79.40$^{\pm \text{0.002}}$ & 82.43 $^{\pm \text{0.005}}$\\
\midrule
\multicolumn{2}{c|}{$\star$\textbf{\textsc{SolAr}: \textit{Ensemble Retrieval and Input}}} & \textbf{79.90} $^{\pm \text{0.001}}$& \textbf{82.44} $^{\pm \text{0.005}}$& \textbf{78.80} $^{\pm \text{0.001}}$& \textbf{82.44} $^{\pm \text{0.005}}$& \textbf{80.99}$^{\pm \text{0.001}}$ & \textbf{82.43} $^{\pm \text{0.005}}$\\
\bottomrule
\end{tabular}
}
\caption{Performance of different retrieval strategies across author percentiles with respect to data size. ``Contro'' denotes controversial situations. \textsc{SolAr}, which ensembles different strategies and input types for controversial and non-controversial situations at test time, achieves the best overall performance. The performance boost is more significant for the bottom 50\% redditors and in controversial situations.}
\label{tab:experimental-results-rag}
\end{table*}

\subsection{Baseline Models}
In addition to our proposed framework using LLMs for reasoning, we implement trainable language models varying in structures and inputs to show the difficulties in understanding individual-level subjectivity. We use pre-trained language models varying in encoder and decoder structures and fine-tune the models with our dataset. We randomly split each redditor's instances into 60/10/30$\%$ to obtain the training, validation, and test set, and run 5-fold cross-validation. Language models are fine-tuned for each redditor, thus we fine-tune 100 distinct models for each model structure.

We implement three encoder-only models, DistilBERT \citep{sanh2019distilbert}, RoBERTa \citep{liu2019roberta}, and DeBERTa-v3 \citep{he2021debertav3}, where the input for each model is a text description of the situation and the output is the redditor's acceptability judgments, 0 or 1. We denote this approach as \textit{Endoer-only Models}.

Denoted as \textit{Endoer-Decoder Models}, we fine-tune encoder-decoder language models, mainly BART \citep{lewis2019bart} and FLAN-T5 \citep{chung2024scaling}. We report the experimental results of these models with the same classification setting as the encoder-only models; models get situation text as input and output either 0 or 1. 

To learn more about each redditor's subjectivity, we test another variation. The models get the same input, but their objective is to generate the redditors' likely reactions (i.e., comments and judgments). While encoder-only models only observe each redditor's binary judgment patterns with respect to the situations, the models fine-tuned with this approach have access to the comments that are authored by the redditors. This approach is denoted as \textit{Endoer-Decoder Models; Seq2Seq}. Table \ref{tab:experimental-results-baseline} illustrates the results of the baseline model. Further details of the baseline model implementation are described in Appendix \ref{sec:appendix-baseline}. 

% Finally, we implement hierarchical models that combine language generation abilities of encoder-decoder models and classification abilities of encoder-only models. In this setup, encoder-decoder models generate each redditor's likely reactions without judgments, and a separate encoder-only model learns likely judgments from the generated comments. 
% For the hierarchical models, we fine-tune encoder-decoder models with two different objectives, the redditor's comments and the abstract values. 

% Since the number of training instances for each redditor might not be sufficient, we try techniques that help fine-tuning the models with smaller number of instances such as LoRA \citep{hu2021lora}. We report the best combinations resulting for each model structure in Table \ref{tab:experimental-results}.

\subsection{RAG-based Models}
We compute the performance for each redditor separately, using the same test set as the ones used in baseline models. As described in \ref{sec:subjective-ground-retrieval}, we represent the individual subjectivity in a several different ways. 

First, we try different retrieval strategies by querying past history data with situations and abstract values that are annotated by LLMs. For querying with abstract values, we report the results of using Schwartz's values and value trade-offs. After retrieval, we differentiate LLM inputs (i.e. few-shot examples) by adding Schwartz's values and value trade-offs that are observed from each redditor's past history. 

Our proposed framework, \textsc{SolAr}, ensembles retrieval strategies based on the difficulty of the test instance. When the test situation is given, \textsc{SolAr} computes its difficulty based on how controversial it is---defined as having less than 70\% agreement among all redditors' judgments\footnote{This is the threshold that \texttt{r/AITAFiltered} uses to identify controversial situations from \texttt{r/AmITheAsshole}.}. For controversial situations, \textsc{SolAr} employs the value-aware retrieval function $R_{val}( \ )$, while it uses the situation-based retrieval function $R( \ )$ for non-controversial cases. In both scenarios, few-shot examples include both the comment and the corresponding value trade-offs.

We use GPT-4.1 \citep{openai2025gpt41} to predict the acceptability judgments of individuals. Each method retrieves five few-shot examples and prompts the LLM twice to estimate a confidence interval. Table \ref{tab:experimental-results-rag} shows the performance of each method. Appendix \ref{sec:appendix-rag} explains more detailed experimental settings for RAG-based models. 

% To compare the effectiveness of subjective ground retrieval, we prompt the LLM with random few-shot examples, where the examples are neither relevant to the test situation nor from the same redditor. 

% \subsection{Reasoning module}
% Reasoning about value conflicts works the best!

\begin{figure}
    \centering
    \includegraphics[width=0.49\textwidth]{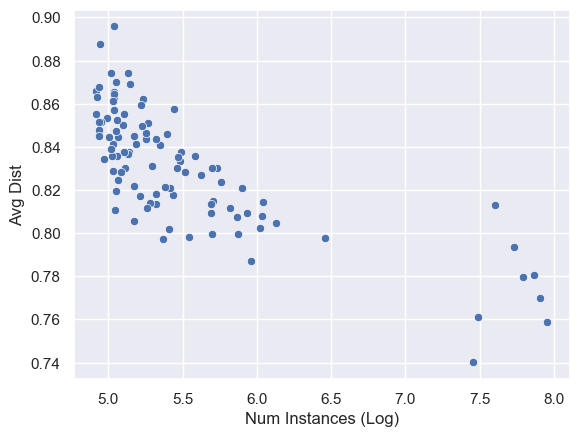}
    \caption{Average distance between the test situation and retrieved instances for each redditor. $x$-axis is the log-scaled numbers of each redditor's data size. When the redditor has more instances, it is more likely that they have commented on situations that are more similar to the test situation in the past.}
    \label{fig:avgdist-numinst}
\end{figure}

\section{Discussion}
In this section, we analyze the inference performance of different models and discuss the effectiveness of each model component. In addition, we perform qualitative analyses of each redditor's subjective ground with respect to annotated abstract values and assess how the proposed framework explains individual-level subjectivity.

\subsection{Inference Performance}
The overall F1 scores of the baseline models in Table \ref{tab:experimental-results-baseline} show that fine-tuning redditors' decision patterns with language models does not solve the problem. The models show fairly good performance only for the redditors who have enough amount of training instances, while they fail to learn the subjectivity of redditors who have a small number of instances and highly skewed judgment patterns. We visualize fine-tuned models' performance with respect to the number of instances and judgment skewness in Appendix \ref{sec:appendix-baseline-performance}.

We also observe that as the model becomes more complex and parameterized, the inference performance worsens---the F1 scores of encoder-decoder models are worse than encoder-only models, and sequence-to-sequence models are even worse than that. This implies that the nature of data scarcity in subjectivity analysis also makes it more difficult to fine-tune a specific individual's perspectives to characterize their subjectivity.

On the other hand, RAG-based models that select the most relevant instances to a test situation work generally well. This answers one of our research questions, \textit{``Can off-the-shelf LLMs account for individual's subjective preference?''}. As RAG-based inference does not require training, the performance of redditors with less amount of instances (i.e. Bottom 50\%) matches the overall F1 scores.

\begin{figure*}[t!]
    \centering
    \includegraphics[width=0.8\textwidth]{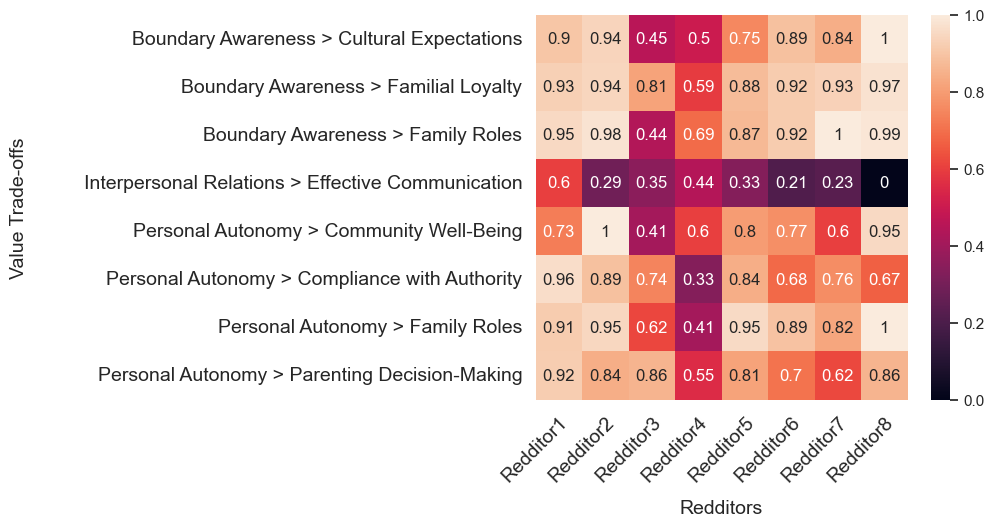}
    \caption{\textbf{\textit{Visualization of subjective preferences.}} The heatmap shows the most common value conflicts among the 8 most active redditors. Each value trade-offs are represented in a ``A > B'' form, and each cell's value means the win rate of value A over value B for a specific redditor. In most cases, redditors' value trade-off systems work differently.}
    \label{fig:val-tradeoff-heatmap}
\end{figure*}

Different retrieval methods and inputs show distinct advantages. First of all, when we compare different inputs within the standard retrieval strategy (i.e. Situation), adding values improves the performance for redditors who have less amount of instances; using comment and value trade-offs improves the F1 score by 2.69\% for the bottom 50\% redditors. Figure \ref{fig:avgdist-numinst} shows that the average distance between the test situation and the retrieved instances increases for redditors with fewer instances. The performance boost we get for the bottom 50\% redditors implies that adding abstract values helps characterize the redditors' subjectivity when the retrieved instances are topically less similar to the test situation.

Abstract values are also helpful when we use them for retrieval. The performance in controversial situations improves when the most relevant instances are retrieved using either Schwartz's values or value trade-offs. This suggests that predicting individuals' subjective preferences benefits from knowing their high-level value system, especially when the test situation is controversial and LLM's own morality does not align well with humans.

\subsection{Diverse Value Trade-offs among Redditors}
A key research hypothesis of this study is that each redditor actively participating in \texttt{r/AmITheAsshole} exhibits distinct subjectivity patterns, making the analysis of individual perspectives a fundamentally different NLP challenge compared to social commonsense reasoning. We validate this hypothesis by visualizing the value trade-off patterns of the 8 most active redditors.

Figure \ref{fig:val-tradeoff-heatmap} displays the 8 most common value trade-offs of the most active redditors. Each cell in the heatmap refers to win rates. For example, $0.9$ in the cell in row 1, column 1 means that among all situations where a value \textit{``Boundary Awareness''} conflicts to \textit{``Cultural Expectations''}, redditor 1 chooses the former over the latter 90\% of the times. For the same situations, redditor 3, who has a $0.45$ win rate, chooses \textit{``Boundary Awareness''} only 45\% of the time, implying they prefer \textit{``Cultural Expectations''} slightly more.

The heatmap tells us that there are some values commonly cherished by most of the redditors. The bottom row of the heatmap, for instance, shows that all redditors prefer \textit{``Personal Autonomy''} over \textit{``Parenting Decision-Making''} in most cases. At the same time, redditors show diverse patterns when it comes to \textit{``Personal Autonomy''} conflicts with other values. For redditor 8, specifically, this value is highly prioritized in most cases, unless it conflicts with \textit{ ``Compliance with Authority''}. This visualization not only suggests that active redditors in \texttt{r/AmITheAsshole} show distinct subjectivity, but explains their value preferences and judgments.

\section{Related Studies}
The closest neighbor of this research is individual subjectivity analysis. \citet{lee2022towards} analyzes the same community, \texttt{r/AmITheAsshole} and learned subjective preference of individuals by computing attention weights between situations and social norms. \citet{plepi2022unifying} used the authorship information with text embeddings and perform the YTA/NTA classification of the redditors in \texttt{r/AmITheAsshole}. In the political framing and agenda-setting domain, \citet{roy-goldwasser-2021-analysis} utilized Moral Foundations theory to characterize real world politicians varying in topics.
%Representing individual-level perspectives is also done in debate setting; \citet{li2018structured} used graph embeddings to associate opnions and individuals together.
More recently, researchers utilized LLMs to ground their generations to a desired persona or personality in many domains, including question answering \citep{zheng-etal-2024-helpful}, interactive simulacra \citep{park2023generative}, and solving causal inference and moral dilemma \citep{nie2023moca}. \citet{choipicle} proposed Persona In-Context Learning which uses Bayesian inference to select the optimal set of persona for a given task. Researchers have also tried providing more direct signals in the prompt using demographic information \citep{durmus2023towards}. 

Incorporating value theories into LLMs is an active research area these days. \citet{van2023differences} analyzed (dis-)agreements between the users by identifying value profiles with Schwartz's value theories. \citet{sorensen2025value} used LLMs as proxy humans with different attributes (e.g. demographic traits, value profiles), and analyzed their distinct behaviors on controversial issues. \citet{bhatia2025computational} used LLMs to characterize individuals’ choices by generating descriptions of benefits and costs of their choices. \citet{sorensen2024value} used concrete examples (i.e. text document) and prompted LLMs to generate corresponding values, rights, and duties. \citet{ye2025measuring} parsed free-form input text into various perceptions and the corresponding Schwartz values by fine-tuning LLMs.

The main differences between our approach and the existing studies in subjectivity analysis and human value analyses with LLMs are; (1) we analyze actual redditors in real world rather than using LLMs as proxy humans, (2) our approach applies more fine-grained buckets for subjectivity, and (3) the novel value trade-off analysis provides additional explanation for individuals' choices. 

Another line of related studies is the Retrieval-Augmented Generation. RAG retrieves the most relevant information to mainly fine-tune language models or help off-the-shelf LLMs infer better on many downstream tasks such as QA \citep{shi2023replug, xu2023retrieval, wang2024knowledge}, reasoning and language understanding \citep{yu2023augmentation, lin2023ra, zhang2023iag}, text summarization and generation \citep{guo2023prompt, jiang2023active, yan2024corrective}, etc. Researchers have also studied whether RAG can be applied to more personal use cases such as recommendation system \citep{rajput2023recommender} and personalized dialog generation \citep{wang2023large, wang2024unims}. These approaches only consider factual information (e.g. purchase history, where the person went two days ago) as a personalized aspect, while our research characterizes subjective perspectives and applies RAG for performance improvement. 

\section{Conclusion}
In this paper, we propose a framework, \textsc{SolAr}, that takes redditors' past comments into account and characterizes their subjective ground using value abstraction. Empirical results show that LLMs can be efficiently used to account for the subjective preference of individuals compared to traditional methods that require fine-tuning. We also show that the performance, especially for redditors with small data sizes and in controversial situations, is improved by retrieving the most relevant instances using abstract values. Furthermore, \textsc{SolAr} provides additional explanations about each redditor's distinct value preference patterns which could later be used to justify LLM inference.

\section*{Limitations}
Although it is not feasible to model the subjectivity of individuals that is perfectly correct and inclusive, representing it with their past comments is still an oversimplified definition of subjectivity. First, past comments in the same community might not cover all aspects of the subjectivity. In future studies, we plan to incorporate more redditor-related information such as their community membership (i.e., what other subreddits they are actively participating in) and activities in other communities to better characterize individuals. Another oversimplification is that it assumes the redditors would judge the situations consistently over time. It would be an interesting direction to analyze whether there are value shifts over time for the redditors.

Another limitation of this study is that the datasets and the tasks are tested only on a specific subreddit. Although \texttt{r/AmITheAsshole} is a huge online community that covers a wide range of situations exhibiting diverse perspectives, the usefulness of our framework will be more strongly validated when we apply our approach to other communities that require subjective perspectives. 

In terms of downstream tasks and the model's performance, our proposed framework shows sub-optimal performance in predicting the correct judgment. Although our framework shows higher improvements when considering controversial situations, our goal is to make LLMs perform well not only in controversial situations but also in other situations.

Lastly, more validations of the generated abstract values are needed. As we use LLMs to freely generate value trade-offs of redditors that are observed from their comments, evaluating the soundness and quality of the values would make the subjectivity representation more powerful and useful. We further plan to validate this process with actual human evaluations and see if LLM-generated values can characterize human values reasonably well. The best way to evaluate LLMs' characterization of individuals would be directly ask it to the target individuals. We leave recruiting human participants to accurately evaluate LLM generations as future work.

\section*{Acknowledgements}
This project was partially funded by  NSF IIS-2048001 and DARPA CCU program. The views are the authors' and should not be interpreted as necessarily representing the official policies, either expressed or implied, of DARPA, or the U.S. Government.

\section*{Ethics Statement}
To the best of our knowledge, this work has not violated any code of ethics. All redditor information is anonymized in this paper as well as in the datasets we share to the public. The redditors are selected purely based on how active they participate in the community thus there is no discrimination in choosing redditors of interest. We denote that this paper poses potential risks where LLMs could misrepresent the subjectivity of individuals by referring to a limited number of past history and making hypotheses on their likely reactions to an unseen situations. We provide the code and datasets for future reproduction.

\bibliography{custom}

\appendix

\section{Data Distribution}
\label{sec:appendix-data-dist}

\begin{figure}[h]
    \centering
    \includegraphics[width=0.49\textwidth]{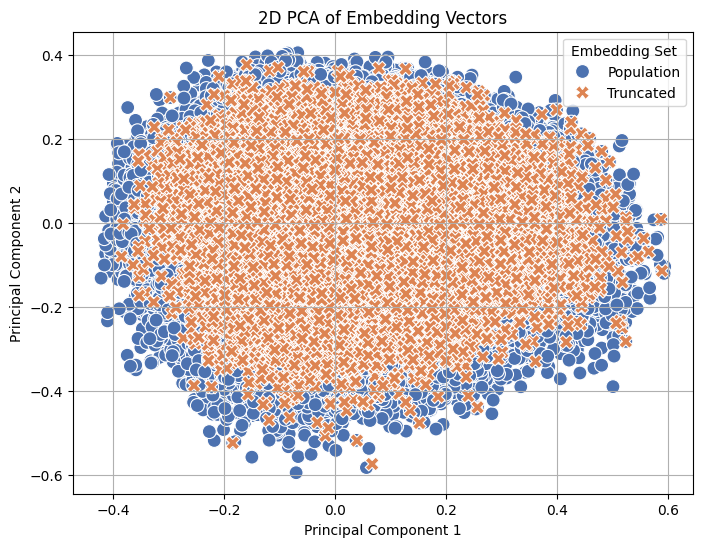}
    \caption{PCA analysis of situation representations of all \texttt{r/AmITheAsshole} posts and truncated posts we use for training and inference. Truncated situations show similar distributions to the population distribution.}
    \label{fig:pca-population-truncated}
\end{figure}

As described in \ref{sec:data-crawl}, we crawl all posts in \texttt{r/AmITheAsshole} from November 2014 to June 2023 and filter out threaded comments. As a result, we have around 217K unique situations with 891K instances in total. Fine-tuning models and performing LLM inference on this massive dataset takes up too many resources. Thus we truncate the dataset. We first identify the most active redditors (i.e. redditors who commented the most). Then we identify the redditors who commented more than a certain threshold\footnote{We set this as 2,000}. We get 8 redditors from this process. After this step, we filter out situations that the 8 most active redditors did not leave comments. As a result, 1.7K unique situations are left.

\begin{figure}[t]
  \centering
  \begin{subfigure}[b]{0.45\textwidth}
    \includegraphics[width=\linewidth]{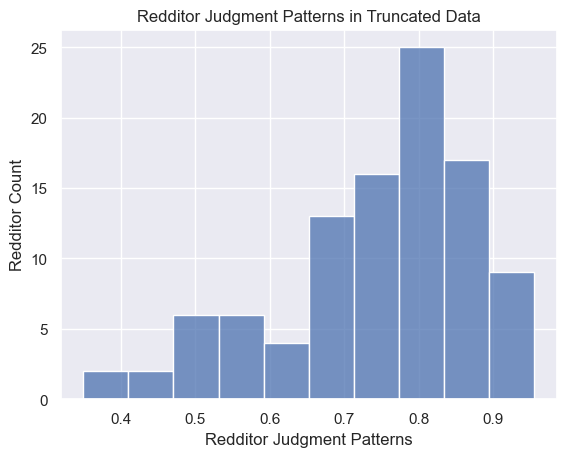}
    % \caption{}
    \label{fig:fig1}
  \end{subfigure}
  % \hfill
  \vspace{1em}
  \begin{subfigure}[b]{0.45\textwidth}
    \includegraphics[width=\linewidth]{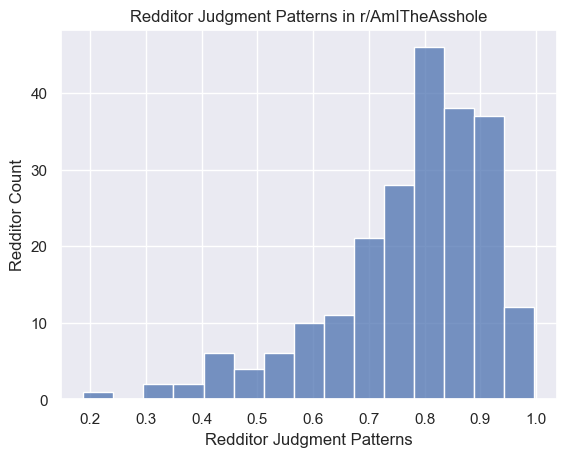}
    % \caption{Caption for figure 2}
    \label{fig:fig2}
  \end{subfigure}
  \caption{Redditors' judgment patterns with respect to the redditor counts in the truncated data (top), and in \texttt{r/AmITheAsshole} (bottom).}
  \label{fig:combined}
\end{figure}

To ensure that the truncated situations and all redditors who commented on these situations are not too different from the population distribution, we first visualize situation representations in a 2-D embedding space using Principal Component Analysis. Figure \ref{fig:pca-population-truncated} visualizes the vector representations of all 217K situations in \texttt{r/AmITheAsshole} (blue dots), and vectors of 1.7K situations in the truncated dataset (orange crosses). The plot implies that the situations in the truncated dataset are not deviated from the population distribution.

We compare the statistics of redditors in \texttt{r/AmITheAsshole} and the truncated dataset. Figure \ref{fig:combined} shows the redditors' judgment patterns (i.e. acceptable or not acceptable) in the truncated data and in all \texttt{r/AmITheAsshole}. A judgment pattern of 1 means that the redditor judged all situations as ``acceptable'', and 0 means they judged all as ``not acceptable''. When the value is 0.5, the redditor has a perfectly balanced judgment history. In the truncated data, redditors tend to judge more situations as ``acceptable'', and the trend is the same in the \texttt{r/AmITheAsshole}. This shows that the redditors in the truncated dataset do not diverge from the overall population.

\section{Value Abstraction}
\subsection{Schwartz's Theory of Basic Human Values}
\label{sec:schwartz-theory}
\citet{schwartz1992universals} defines theory of basic human values as:
\begin{itemize}[noitemsep, topsep=0pt]
    \item Self-direction: ``independent thought and action---choosing, creating, and exploring''
    \item Stimulation: ``excitement, novelty and challenge in life''
    \item Hedonism: ``pleasure or sensuous gratification for oneself''
    \item Achievement: ``personal success through demonstrating competence''
    \item Power: ``social status and prestige, control or dominance over people and resources''
    \item Security: ``safety, harmony, and stability of society, of relationships, and of self''
    \item Conformity: ``restraint of actions likely to upset or harm others and violate social norms''
    \item Tradition: ``respect of the customs and ideas that one's culture or religion provides''
    \item Benevolence: ``preserving and enhancing the welfare of those within-group)''
    \item Universalism: ``protection for the welfare of all people and for nature''
\end{itemize}
% In a more generalized view of the values, these values can be grouped into four categories, Openness to Change (self-direction, stimulation), Self-Enhancement (hedonism, achievement, power), Conservation (security, conformity, tradition), and Self-Transcendence (benevolence, universialism). Openness to Change and Conservation contradicts to each other, and Self-Enhancement and Self-Transcendence contradicts to each other as well.

In order to annotate situations and comments to these fixed values, we prompt \textit{gpt-4o-0806} model to generate values observed from the text. Below is an example prompt:

\begin{tcolorbox}[title=Prompt for Annotating Schwartz's Values, colback=blue!5, colframe=blue!50, fonttitle=\bfseries\footnotesize, fontupper=\footnotesize]
You will be given a social situation and a Redditor's comment. Your job is to find the top two most salient Schwartz's Basic Human Values that are observed in the situation and in the comment. Follow the format below without generating any other explanation.
\newline\newline
Input Format:\newline
[Situation] situation\_description\newline
[Comment] redditor\_comment\newline
\newline
Output Format:\newline
[Situation] Schwartz\_value\_A, Schwartz\_value\_B\newline
[Comment] Schwartz\_value\_C, Schwartz\_value\_D\newline
\newline
\#\#\#\newline
[Situation] Not babysitting my niece\newline
[Comment] When you talk to your brother ..\newline
\end{tcolorbox}

\subsection{Clustering Value Conflicts and Trade-offs}

Similar to annotating Schwartz's values, we prompt \textit{gpt-4o-0806} model and generate value conflicts and trade-offs. Below is an example prompt:

\begin{tcolorbox}[title=Prompt for Annotating Value Conflicts, colback=blue!5, colframe=blue!50, fonttitle=\bfseries\footnotesize, fontupper=\footnotesize]
You will be given a situation. Your task is to understand the situation, and tell me what kind of moral values are conflicting.
\newline\newline
[Situation] Not babysitting my niece ...\newline\newline
Tell me what kind of moral values are conflicting in the situation. Note that the two conflicting values can't be chosen at the same time (i.e. trade-off), and people's behaviors and attitudes will be different based on which value they choose.\newline 
Highlight one or more core conflicting values that describe the situation the best. Values should be generic; proper nouns or pronouns should not be included. Write values in phrases, not in sentences, and each value should have at least 3 words. Your response should follow the format:\newline
[\newline
    "A vs. B", \newline
    "C vs. D", \newline
    ..\newline
]
\end{tcolorbox}

After the conflicting values are annotated for the situations, we prompt the GPT model again to analyze the comments from each redditor. Below is a prompt:

\begin{tcolorbox}[title=Prompt for Annotating Value Trade-offs, colback=blue!5, colframe=blue!50, fonttitle=\bfseries\footnotesize, fontupper=\footnotesize]
You will be given a [Situation] and a list of [Conflicting Values] observed in the situation. There's Person X who leaves a [Comment]. \newline\newline
[Task] Determine which items in the [Conflicting Values] are mostly related to Person X's [Comment]. You may select only one of them, or select multiple items. Generate conflicting values observed in the comment ONLY IF none of the [Conflicting Values] are related to the [Comment]. \newline
Determine which value is more important to Person X. For instance, "A vs. B" is a chosen conflicting value and Person X thinks A is more important than B in the comment, the answer should be "A > B".\newline
\newline
Your response should follow the format:\newline
[\newline
        "A > B",\newline
        "D > C",\newline
        ...\newline
]\newline
\newline
\#\#\#\newline
[Situation] Not babysitting my niece\newline
[Conflicting Values] Autonomy vs. Family\newline
[Comment] When you talk to your brother ..\newline
\end{tcolorbox}

\label{sec:cluster-value-conflicts}
\begin{figure}
    \centering
    \includegraphics[width=0.49\textwidth]{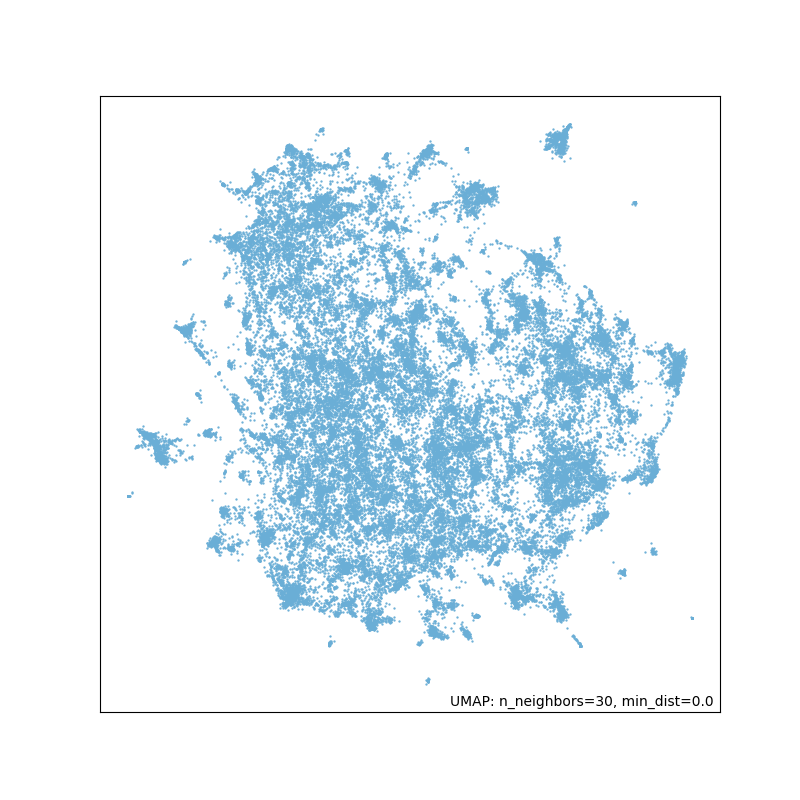}
    \caption{Visualization of Umap embeddings of value representations.}
    \label{fig:umap}
\end{figure}

The general framework of creating clusters from the annotated values follow the approaches suggested by \citet{lam2024concept}. The authors first cluster the texts using HDBSCAN, then expand the concepts by prompting LLMs. The initial clusters are formed using HDBSCAN \citep{mcinnes2017hdbscan}, and we later use LLMs to create additional clusters for values that remain uncategorized in the initial clustering phase, resulting in 111 clusters in total. 

For the generated values that are conflicting in the situations in the dataset, we first obtain their vector representations. We use OpenAI's text-embedding-3-large model \citep{openai2024textembedding3small}, and reduced the dimensions to 256. We then further reduce the dimensionality using umap embeddings \citep{mcinnes2018umap}. Figure \ref{fig:umap} shows the 2-D visualization of umap embedding of all value representations, with number of neighbors as 30 and minimum distance as 0. 

After this step, we use HDBSCAN to generate initial clusters while enforcing minimum of 100 items in each cluster. After obtaining the initial cluster, we gather all uncategorized values. We then compute the distance between each of the uncategorized value and initial cluster representations. If the distance is closer than a threshold (0.95), we assign the item to the cluster. For values that are not close enough to any other clusters, we group them using K-means clustering.

After assigning all value representations to a corresponding cluster, we ask \textit{gpt-4o-mini} model to come up with a summary of unifying themes and patterns. Below is the template for the prompt adopted from \citet{lam2024concept}:

\begin{tcolorbox}[title=Prompt for Expand Concepts, colback=blue!5, colframe=blue!50, fonttitle=\bfseries\footnotesize, fontupper=\footnotesize]
I have this set of bullet point summaries of text examples:\newline
\{examples\}\newline\newline
Please write a summary of unifying patterns for these examples.\newline
For each high-level pattern, write a 5 word NAME for the pattern and an associated one-sentence ChatGPT PROMPT that could take in a new text example and determine whether the relevant pattern applies.\newline
Please also include 3 example\_ids for items that BEST exemplify the pattern.
\end{tcolorbox}

\subsection{Comparison}
\label{sec:appendix-value-comparison}
\begin{figure*}[t!]
    \centering
    \includegraphics[width=0.95\textwidth]{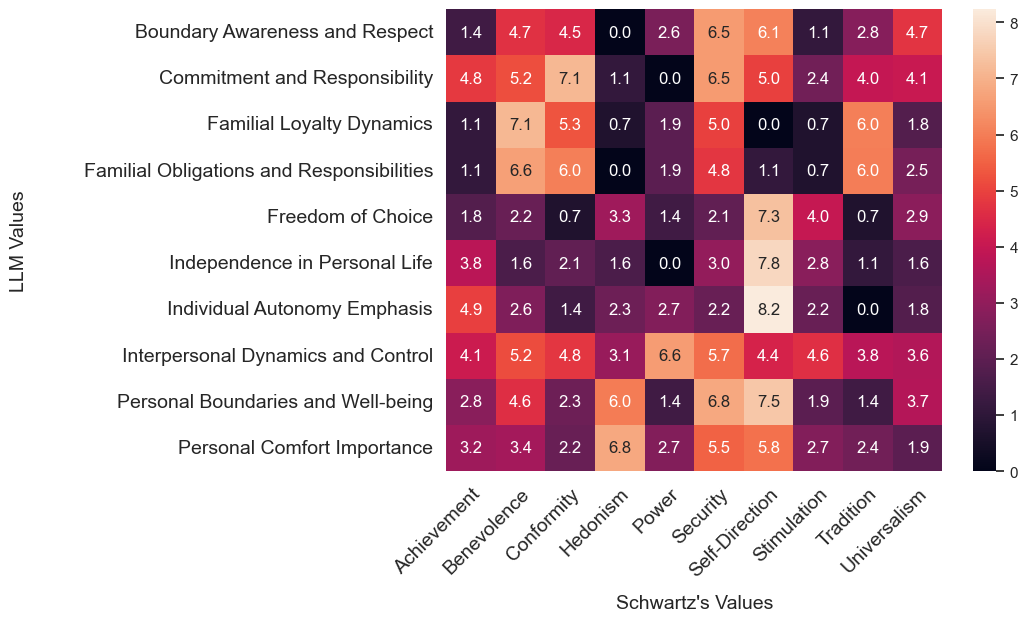}
    \caption{Mapping results of LLM-generated values to Schwartz's values. Numbers in each cell mean the log-sclaed count of the number of co-occurrences.}
    \label{fig:cluster-to-schwartz}
\end{figure*}

In order to analyze the usefulness of LLM-generated values and whether they cover the fixed set of values that are defined in a top-down approach, we map these values to the Schwartz's values. 

Figure \ref{fig:cluster-to-schwartz} shows the mapping results. For each of the situation that has annotations of both LLM-generated values and Schwartz's values, we count the co-occurrences between these values. For instance, if a situation's annotated Schwartz's value is ``\textit{Security}'', and at the same time, its annotated values that conflict is ``\textit{Boundary Awareness and Respect}'', then these two values have a co-occurrence. In the figure, we compute the log-scale of all the co-occurrence counts.

For each of the LLM-generated values, they cover multiple dimensions of Schwartz's values, suggesting that these values have richer context information about the situations. Moreover, in some cases, these LLM-generated values cover conflicting Schwartz's values together. For example, ``\textit{Personal Boundaries and Well-being}'' value has high number of co-occurrences with ``\textit{Security}'' and ``\textit{Self-Direction}''. These two Schwartz's values conflict to each other, thus they are not categorized into the same bucket using the Schwartz's values. This shows that LLM-generated values can be applied in a more flexible way, while preserving meaningful insights from the theories supported by Schwartz's values.

\section{Model Implementation Details}
\subsection{Baseline Model Implementation}
\label{sec:appendix-baseline}
For all fine-tuned language models, we perform hyperparameter searching on training batch size and learning rates. For DistilBERT-base, the best working combination is 16 and 3.962e-5, for RoBERTa-base, it is 32 and 3.97838e-5, for DeBERTa-v3, it is 4 and 7.378218e-5. For encoder-only models, fine-tuning models for each redditor takes approximately 10 to 25 minutes on a single A30 GPU. Fine-tuning encoder-decoder models, BART and FLAN-T5, took around 15 to 30 minutes, respectively. Stretching this to 100 redditors and 5 fold experiments, running each model structure took a day to two days.

\subsection{RAG-based Model Implementation}
\label{sec:appendix-rag}
For inference using GPT-4.1 models, each of the experiment costs around USD 30, where the costs for input query takes about USD 29 and the rest is for the output which is either 0 or 1. This was based on the batch prompting, which is half of the original price.

\section{Model Performance}
\label{sec:appendix-baseline-performance}

\begin{figure*}[t]
  \centering
  \begin{subfigure}[b]{0.47\textwidth}
    \includegraphics[width=\linewidth]{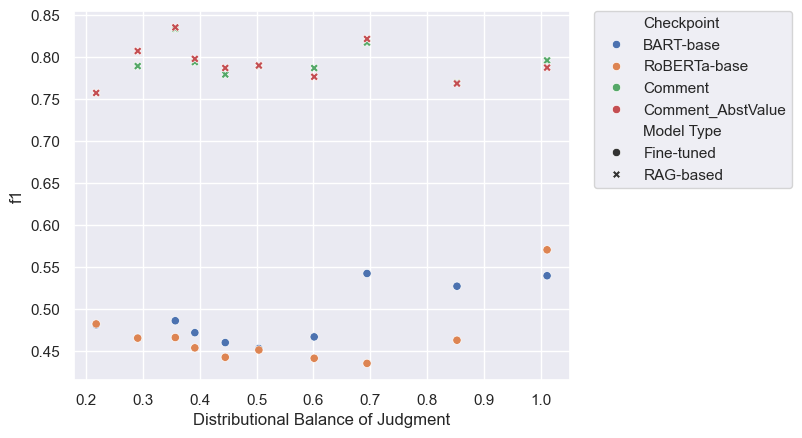}
    % \caption{}
    \label{fig:fig1}
  \end{subfigure}
  % \hfill
  % \vspace{1em}
  \begin{subfigure}[b]{0.47\textwidth}
    \includegraphics[width=\linewidth]{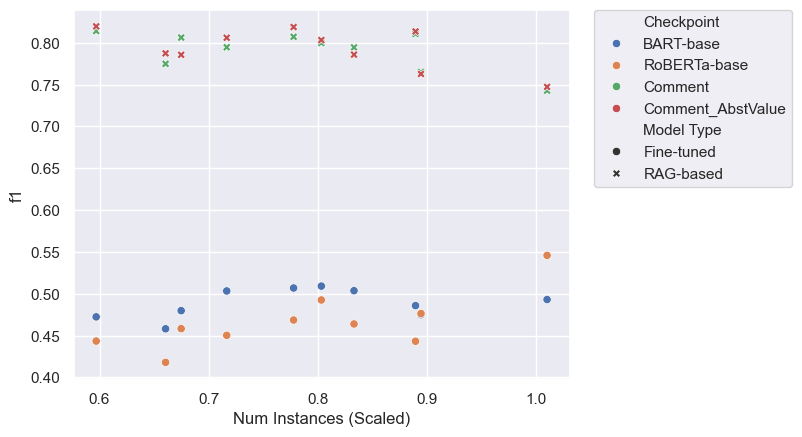}
    % \caption{Caption for figure 2}
    \label{fig:fig2}
  \end{subfigure}
  \caption{Macro F1 score of fine-tuned models and RAG-based models. $x$-axis represents the attribute of redditors. The left figure shows how balanced the redditors' judgment patterns are (more balanced if the value is closer to 1), and the right figure shows how many instances the redditors have.}
  \label{fig:avgdist-combined}
\end{figure*}

Figure \ref{fig:avgdist-combined} illustrates the macro F1 score of the redditors varying in judgment distribution balance and the number of instances. For fine-tuned models, the F1 score decreases as the number of instances decreases and the judgment patterns become more skewed. For the RAG-based models, on the other hand, the F1 socre does not heavily depend on the judgment distributions and the number of instances for each redditor, as the RAG-based models can perform inference based on the few-shot examples only.

% \section{Experimental Results}
% \label{sec:appendix-exp-results}
% In this section, we report the performance of different models and subjective ground representations

\end{document}